\documentclass{article}
\usepackage[main, final]{neurips_2026}
\usepackage[utf8]{inputenc}
\usepackage[T1]{fontenc}
\usepackage{hyperref}
\usepackage{url}
\usepackage{booktabs}
\usepackage{amsfonts}
\usepackage{nicefrac}
\usepackage{microtype}
\usepackage{xcolor}

\usepackage{graphicx}
\usepackage{amsmath}
\usepackage{amssymb}
\usepackage{amsthm}
\usepackage{amsfonts}
\usepackage{amscd}
\usepackage{algorithm}
\usepackage{algpseudocode}
\usepackage{adjustbox}
\usepackage{float}
\usepackage{multirow}
\usepackage{subcaption}
\usepackage[most]{tcolorbox}
\usepackage{xcolor}
\definecolor{promptbg}{RGB}{248,248,248}
\definecolor{promptframe}{RGB}{210,210,210}
\newtcolorbox{promptbox}{
    colback=promptbg,
    colframe=promptframe,
    boxrule=0.5pt,
    arc=2pt,
    left=6pt,
    right=6pt,
    top=6pt,
    bottom=6pt,
    fontupper=\ttfamily\small,
    breakable
}
\title{DriveMind: Mind-Evolving Belief Tracking for \\ Closed-Loop Autonomous Driving}
\author{
  Yan Shi \\
  University of Washington \\
  \texttt{yanshi@uw.edu}
  \And
  Kehua Chen\thanks{Corresponding author.} \\
  University of Washington \\
  \texttt{zeonchen@uw.edu}
  \And
  Jiarui Qi \\
  University of Washington \\
  \texttt{jiaruiqi@uw.edu}
  \AND
  Yihong Tang \\
  McGill University \\
  \texttt{yihong.tang@mail.mcgill.ca}
  \And
  Haicheng Liao \\
  University of Michigan \\
  \texttt{haichenl@umich.edu}
  \And
  Yinhai Wang\footnotemark[1] \\
  University of Washington \\
  \texttt{yinhai@uw.edu}
}
\begin{document}
\maketitle
\begin{abstract}
  Recent multimodal large language models (MLLMs) can perceive a traffic scene, reason about it, and propose a driving action within a single model. However, existing MLLM-based drivers reason from scratch at each frame and commit to a single explanation, discarding earlier hypotheses about the scene. When the scene is ambiguous, the agent locks in on the wrong interpretation and acts on it before evidence accumulates. In this paper, we propose DriveMind, an MLLM-based driving framework that keeps several hypotheses about the scene alive at the same time and updates their weights as new frames arrive. Inspired by particle filtering, DriveMind propagates each hypothesis through a VLM, scores it against the current observation, and resamples to maintain diversity. On Bench2Drive, adding belief tracking to the same driving pipeline improves the Driving Score (DS) from 68.50 to 81.37 and the Success Rate (SR) from 41.36\% to 54.09\%. The largest gains occur on ambiguous scenarios such as unsignalized intersections (+20.0 DS) and oncoming-traffic interactions (+20.5 DS). Beyond the overall metrics, DriveMind also improves fine-grained interactive driving abilities such as merging, emergency braking, and give-way. The code is available at: \url{https://github.com/shiyannacl/drivemind}.
\end{abstract}
\section{Introduction}
\label{sec:intro}
MLLMs have recently been adopted for autonomous driving, where a single model takes raw sensor input and outputs both a textual rationale and a driving decision~\citep{cui2024survey}. This makes them a natural fit for tasks where the model must explain its choices, such as following natural-language navigation commands or justifying a lane change.
\begin{figure}[htbp]
    \centering
    \includegraphics[
    width=.85\linewidth,
    trim=25 0 0 0,
    clip
]{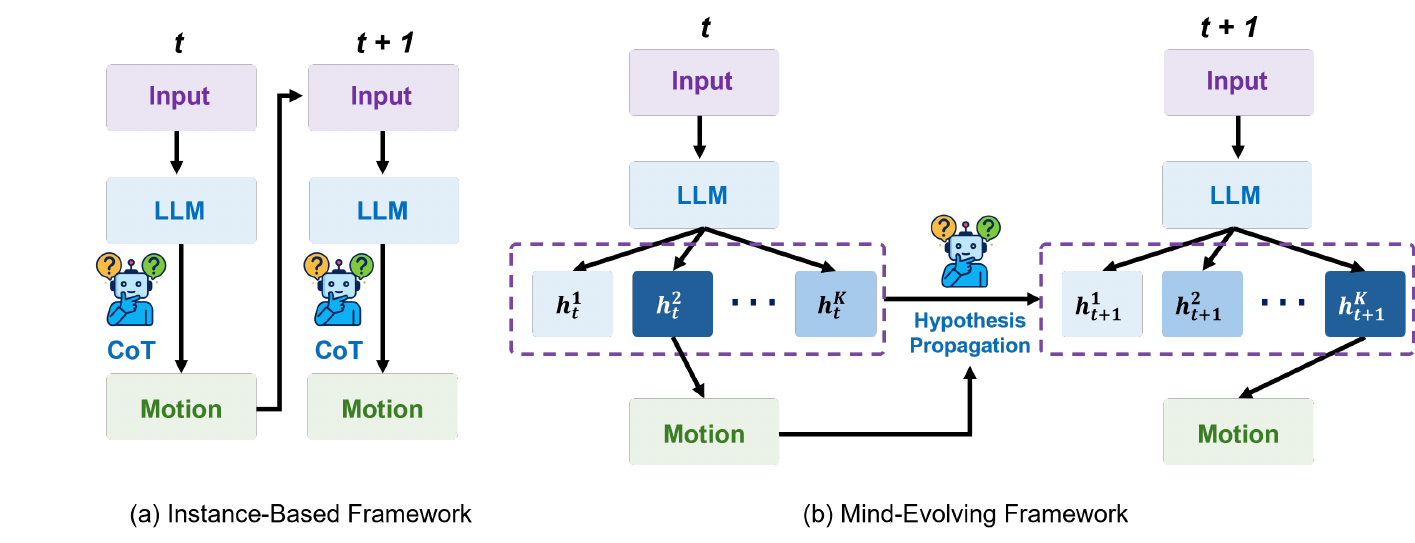}
    \caption{\textbf{From single-chain reasoning to mind-evolving belief.} (a) Instance-Based Framework: one rationale per timestep. (``CoT'' stands for ``Chain of Thought'') (b) Mind-Evolving Framework: multiple hypotheses $\{h_t^k\}$ propagated over time; darker color indicates higher hypothesis weight.}
    \label{fig:first}
\end{figure}
However, driving differs from one-shot question-answering tasks. The agent observes the scene only partially, and the scene keeps changing as it moves. Traffic is also full of ambiguity. For example, a slowing vehicle may be yielding, preparing to turn, or reacting to an unseen obstacle; a pedestrian near the curb may intend to cross or remain waiting. In these cases, multiple explanations can remain plausible at the same time. Several existing MLLM-based driving systems generate a single rationale or planned trajectory at each inference step~\citep{xu2024drivegpt4,mao2023gpt,renz2025simlingo}. This single-chain reasoning is fluent but brittle: the model picks one interpretation too early, then has to act on it whether or not later frames support it. Figure~\ref{fig:first} contrasts the two paradigms: a single chain of thought per frame, versus a set of hypotheses that is carried forward and reweighted over time.
This premature commitment is especially problematic in closed-loop control. Once the model forms a wrong interpretation, the resulting action may change subsequent observations in ways that reinforce the initial mistake.  Existing uncertainty estimation methods for large language models (LLMs) mainly focus on output variability across multiple generations~\citep{ye2024benchmarking,zhang2024luq}. These methods are useful, but they measure uncertainty over multiple answers to the same question, not how uncertainty changes as new frames arrive.
To address this limitation, we propose \textit{DriveMind}, a closed-loop driving framework that maintains a time-evolving belief state over multiple reasoning hypotheses. DriveMind represents the agent's reasoning state as a weighted set of natural-language hypotheses. Each hypothesis describes a possible interpretation of the scene, and its weight represents its relative support under the accumulated observations. As new frames arrive, DriveMind updates the hypotheses themselves and reweights them based on the new evidence.
For real-time closed-loop control, DriveMind adopts an asynchronous fast-slow Vision-Language-Action (VLA) architecture. The slow subsystem uses an MLLM to perform multi-hypothesis reasoning and belief tracking. The fast subsystem runs at control frequency and uses a diffusion-based controller to generate waypoint trajectories conditioned on the latest belief. On Bench2Drive, adding belief tracking improves DS from 68.50 to 81.37 and SR from 41.36\% to 54.09\%. The largest gains occur in scenarios requiring intent disambiguation, such as unsignalized intersections (+20.0 DS). On NAVSIM, belief tracking improves PDMS from 78.2 to 81.0. Ablations support the effectiveness of belief tracking, while runtime analysis demonstrates the efficiency of asynchronous inference.
Our main contributions are summarized as follows:
\begin{itemize}
    \item We introduce \textit{DriveMind}, a closed-loop driving framework that maintains a weighted set of natural-language hypotheses and updates them as new observations arrive.
    \item We design an asynchronous fast-slow VLA architecture where a slow MLLM reasoner performs belief tracking and a fast diffusion controller generates real-time trajectories.
    \item We show on Bench2Drive and NavSim that explicit belief tracking improves driving performance and robustness, with the largest gains on long-tail scenarios requiring intent disambiguation.
\end{itemize}
\section{Related Work}
\label{sec:related}
\subsection{Autonomous Driving with LLM}
LLMs and MLLMs have been applied to autonomous driving across a range of tasks, from scene understanding and decision-making to motion planning. Existing systems such as LLM-Driver~\citep{drivellm}, DriveGPT4~\citep{xu2024drivegpt4}, DriveLM~\citep{sima2024drivelm}, GPT-Driver~\citep{mao2023gpt}, and LanguageMPC~\citep{sha2025languagempc} use language models for scene understanding, decision making, or motion planning. Other works incorporate commonsense knowledge or linguistic feedback to improve interpretability and adaptability~\citep{fu2024drive,wen2023dilu,feedback}. More recent closed-loop frameworks, including LMDrive~\citep{shao2024lmdrive}, SimLingo~\citep{renz2025simlingo}, and ReCogDrive~\citep{li2025recogdrive}, further connect vision-language reasoning with autonomous driving control. These approaches cover a range of reasoning and memory mechanisms; for example, DiLu includes a memory component and reflection to accumulate driving experience~\citep{wen2023dilu}. DriveMind focuses on carrying forward and reweighting multiple competing scene interpretations as new observations arrive.
\subsection{Uncertainty in Reasoning and Prediction}
Uncertainty estimation for LLMs has been studied from several perspectives. Some methods use step-level uncertainty signals or chain-of-thought information to improve reasoning or response-wise uncertainty estimates~\citep{yin2024reasoning,zhang2025cot}. Other methods assess reasoning quality and output uncertainty using graph edit distance, coherence scoring, or NLI-based decomposition~\citep{da2025understanding,golovneva2022roscoe,lin2023generating,zhang2024luq}. These methods provide useful estimates of output uncertainty, but they mainly focus on static responses or final predictions rather than how uncertainty evolves over time during closed-loop interaction.
Classical planners address this temporal dimension through explicit belief maintenance over predefined state spaces~\citep{cunningham2015mpdm,zhang2020eudm,ding2021epsilon}, but predefined state and behavior representations can limit coverage of rare or open-ended scenarios. DriveMind borrows the idea of temporal belief propagation and uses a VLM to generate hypotheses, allowing hypotheses to express scene interpretations in natural language.
\begin{figure*}[htbp]
    \centering
    \includegraphics[width=\textwidth]{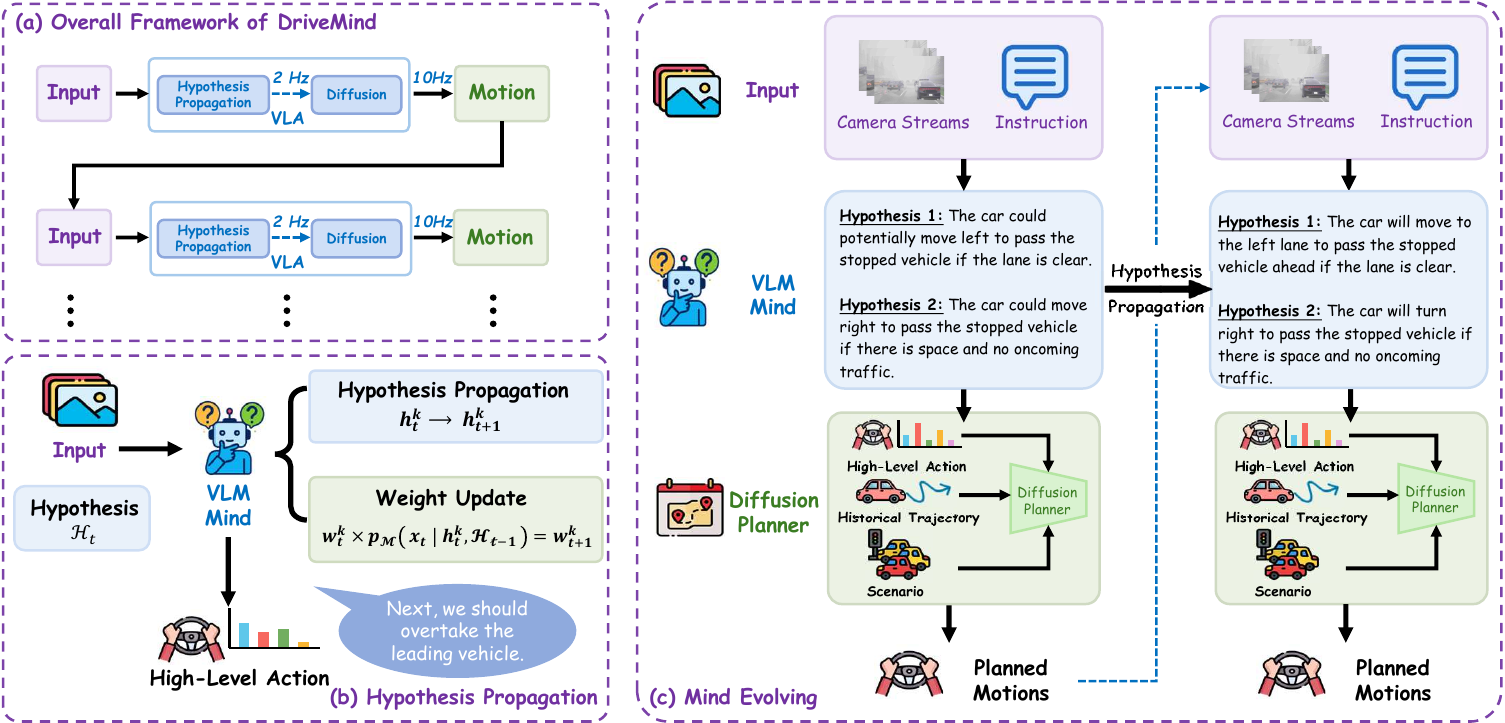}
    \caption{\textbf{\textit{DriveMind} overview.}
    (a) An asynchronous fast--slow architecture: a slow vision-language model (VLM) maintains a multi-hypothesis belief state, while a fast diffusion planner generates trajectories.
    (b) Belief tracking through hypothesis propagation and reweighting.
    (c) The belief state evolves as new observations arrive.}
    \label{fig:overview}
\end{figure*}
\section{Problem Formulation}
\label{sec:formulation}
We formulate autonomous driving as a sequential decision-making process in an interactive environment $\mathcal{E}$. At each time step $t$, the agent receives an observation $\boldsymbol{x}_t$, executes an action $a_t$, and the environment transitions to $\boldsymbol{x}_{t+1} = \mathcal{E}(\boldsymbol{x}_t, a_t)$. Let
\begin{equation}
\mathcal{H}_t = (\boldsymbol{x}_{1:t}, a_{1:t-1})
\end{equation}
denote the history available at time $t$.
Let $h_t \in \mathcal{R}$ denote a latent reasoning hypothesis, where $\mathcal{R}$ is the open-ended space of natural-language interpretations of the scene. A VLM-based driving policy can then be written as a marginalization over hypotheses:
\begin{equation}
p(a_t \mid \mathcal{H}_t)
=
\sum_{h_t \in \mathcal{R}}
p(a_t \mid h_t, \mathcal{H}_t)
p(h_t \mid \mathcal{H}_t).
\label{eq:hypothesis_marginal}
\end{equation}
The key challenge is that the reasoning space $\mathcal{R}$ is open-ended and cannot be enumerated explicitly. A single-hypothesis driving policy can be viewed as approximating Eq.~\eqref{eq:hypothesis_marginal} by selecting one hypothesis and conditioning action generation on it. However, this approximation forces an early commitment: if the chosen hypothesis turns out to be wrong, the agent has already executed an action based on it, and the resulting observations may make recovery difficult.
DriveMind instead maintains a finite approximation to the posterior over reasoning hypotheses. At timestep $t$, we represent the belief state as a weighted set of $K$ natural-language hypotheses:
\begin{equation}
\mathcal{B}_t
=
\{(h_t^k, w_t^k)\}_{k=1}^{K},
\qquad
\sum_{k=1}^{K} w_t^k = 1,
\label{eq:belief_state}
\end{equation}
where $h_t^k$ is the $k$-th natural-language hypothesis and $w_t^k$ is its normalized belief weight. This weighted set gives a discrete approximation to the posterior:
\begin{equation}
p(h_t \mid \mathcal{H}_t)
\approx
\sum_{k=1}^{K}
w_t^k \delta(h_t = h_t^k),
\label{eq:posterior_approx}
\end{equation}
where $\delta(\cdot)$ denotes a point mass on a hypothesis.
Substituting this approximation into Eq.~\eqref{eq:hypothesis_marginal} yields:
\begin{equation}
p(a_t \mid \mathcal{H}_t)
\approx
\sum_{k=1}^{K}
w_t^k
p(a_t \mid h_t^k, \mathcal{H}_t).
\label{eq:action_approx}
\end{equation}
The belief state $\mathcal{B}_t$ is propagated through the closed-loop process and also serves as the conditioning signal for trajectory generation.
\section{Methodology}
\label{sec:method}
As illustrated in Figure~\ref{fig:overview}, DriveMind has two subsystems running asynchronously. A slow VLM reasoner updates the multi-hypothesis belief state $\mathcal{B}_t$ at a low frequency (around 2~Hz). A fast diffusion controller runs at control frequency (around 10~Hz) and generates waypoint trajectories conditioned on the most recently available belief $\mathcal{B}_t$; between two slow updates, the controller reuses the last belief without waiting. We describe the slow subsystem in Section~\ref{sec:propagation} and the fast subsystem in Section~\ref{sec:control}.
\subsection{High-Level Reasoner: Multi-Hypothesis Belief-State Propagation}
Given the previous belief state $\mathcal{B}_{t-1}=\{(h_{t-1}^k,w_{t-1}^k)\}_{k=1}^{K}$, the high-level reasoner updates both the hypotheses and their weights using the new observation $\boldsymbol{x}_t$. We use a particle-filter-style~\citep{gordon1993novel,doucet2001sequential} update with three steps: predict,
correct, and resample, where each hypothesis $h_t^k$ plays the
role of a particle and its weight $w_t^k$ reflects its current
plausibility.
\paragraph{Hypothesis Propagation (Predict).}
\label{sec:propagation}
For each previous hypothesis $h_{t-1}^k$, the VLM samples an updated hypothesis conditioned on the previous hypothesis, the previous action, and the current observation:
\begin{equation}
h_t^k
\sim
p_{\mathcal{M}}
(\cdot \mid h_{t-1}^k, a_{t-1}, \boldsymbol{x}_t, \mathcal{H}_{t-1}).
\label{eq:propagate}
\end{equation}
This equation describes a prompted VLM update inspired by particle filtering, without specifying an explicit probabilistic state-transition model.
\paragraph{Evidence-Consistency Reweighting (Correct).}
After hypothesis propagation, we then ask the VLM to score how well each updated hypothesis is supported by the current observation:
\begin{equation}
s_t^k = f_{\mathcal{M}}(h_t^k, \boldsymbol{x}_t, \mathcal{H}_{t-1}),
\label{eq:score}
\end{equation}
where $f_{\mathcal{M}}(\cdot)$ is a VLM-based scoring function. In our implementation, we prompt the VLM with the hypothesis text, the current observation, and a fixed scoring rubric; the VLM outputs an integer score in [1,10] which we rescale to [0,1]. For example, a hypothesis that a pedestrian is likely to cross should receive a higher score when the pedestrian moves toward the lane and a lower score when the pedestrian remains stationary. We provide the prompt template in Appendix~\ref{app:prompt1}.
We update each hypothesis weight by combining its previous weight with an exponential factor based on its consistency score:
\begin{equation}
\tilde{w}_t^k = w_{t-1}^k \exp(s_t^k / \beta),
\label{eq:weight_unnorm}
\end{equation}
\begin{equation}
w_t^k = \frac{\tilde{w}_t^k}{\sum_{j=1}^{K} \tilde{w}_t^j},
\label{eq:weight}
\end{equation}
Here, $\beta>0$ controls the sensitivity to score differences, and the normalized weights represent relative support among hypotheses rather than calibrated probabilities.
\paragraph{Diversity Maintenance (Resample).}
A common issue in particle filters is degeneracy, where one hypothesis captures all the weight. To maintain diversity, we monitor the effective sample size (ESS)~\citep{kim2025hypothesis}:
\begin{equation}
ESS_t = \frac{1}{\sum_{k=1}^K (w_t^k)^2}.
\label{eq:ESS}
\end{equation}
When $ESS_t < \alpha K$, where $\alpha$ is a predefined threshold, we resample $K$ hypotheses from the current belief state according to their weights. Because naive resampling can produce near-duplicate hypotheses, we add a diversity-aware regeneration step. We embed each hypothesis with the VLM's text encoder and compute pairwise
cosine similarity. For pairs above a threshold $\rho$ (we use $\rho=0.9$), we keep the higher-weighted hypothesis and prompt the VLM to regenerate the other so that it remains consistent with
$\boldsymbol{x}_t$ but covers a different plausible interpretation. See Appendix~\ref{app:diversity}.
\paragraph{Maneuver Marginal.} In addition to maintaining the belief state, the VLM also predicts a high-level maneuver distribution for each hypothesis, $p_{\mathcal{M}}(u_t \mid h_t^k, \mathcal{H}_t)$, where $u_t$ is a discrete action label defined in Appendix~\ref{app:maneuver}. Marginalizing over the belief gives an interpretable maneuver distribution:
\begin{equation}
p(u_t \mid \mathcal{H}_t) = \sum_{k=1}^{K} w_t^k \,
p_{\mathcal{M}}(u_t \mid h_t^k, \mathcal{H}_t),
\label{eq:maneuver_marginal}
\end{equation}
which we use for analysis and visualization (Figure~\ref{fig:case}). The continuous trajectory itself is generated by the diffusion controller in Section~\ref{sec:control}, conditioned on the belief context $c_t$ rather than on $u_t$.
\subsection{Low-Level Controller: Belief-Conditioned Trajectory Generation}
\label{sec:control}
We use a diffusion-based controller $\mathcal{D}$ to map the latest belief into a short-horizon trajectory. Rather than conditioning on the single top-ranked hypothesis, we feed the controller a weighted aggregation of all hypotheses, so that uncertainty in the belief naturally translates into smoother conditioning when no single hypothesis dominates.
\paragraph{Belief Embedding Aggregation.}
To condition the controller on $\mathcal{B}_t=\{(h_t^k,w_t^k)\}_{k=1}^K$, we encode each hypothesis with the VLM's text encoder and take the mean-pooled last-layer hidden state, denoted $E_{\mathcal{M}}(h_t^k)$. The belief context is the weight-aggregated embedding:
\begin{equation}
c_t = \sum_{k=1}^{K} w_t^k E_{\mathcal{M}}(h_t^k).
\label{eq:belief_context}
\end{equation}
This representation combines information from all hypotheses according to their belief weights.
\paragraph{Diffusion-Based Control Policy.}
We model motion generation as conditional denoising~\citep{ho2020denoising,chi2025diffusion}. The controller $\mathcal{D}$ outputs a short-horizon waypoint sequence $a_t = [p_{t,1}, \ldots, p_{t,H}]$ with $H = 8$, where each waypoint $p_{t,i} \in \mathbb{R}^2$ is the $(x,y)$ offset in the ego frame. At inference time, we initialize the trajectory from Gaussian noise $a_t^{T} \sim \mathcal{N}(0, I)$ and iteratively apply the reverse denoising process.
Given a noisy motion $a_t^\tau$ at diffusion step $\tau$, the noise predictor $\epsilon_\theta$ is conditioned on the belief context $c_t$ and visual evidence $\Phi(\mathcal{H}_t)$:
\begin{equation}
\hat{\epsilon} = \epsilon_\theta(a_t^\tau, \tau, c_t, \Phi(\mathcal{H}_t)).
\label{eq:noise_predictor}
\end{equation}
The predicted noise $\hat{\epsilon}$ is used to update $a_t^\tau$ toward a cleaner sample $a_t^{\tau-1}$. After the reverse steps, we obtain the denoised trajectory $a_t = a_t^0$, which is executed in a receding-horizon manner.
DriveMind is trained in two stages. In the first stage, the high-level reasoner is trained with maneuver supervision and score prediction supervision:
\begin{equation}
\mathcal{L}_{\text{high}} =
\mathcal{L}_{\text{man}} +
\lambda_{\text{score}} \mathcal{L}_{\text{score}} .
\end{equation}
\paragraph{Maneuver Prediction Loss.}
Given the belief state $\mathcal{B}_t$, we marginalize over hypotheses to obtain the maneuver distribution and apply negative log-likelihood against the ground-truth maneuver $u_t^*$:
\begin{equation}
\mathcal{L}_{\text{man}}
=
-\log \left(
\sum_{k=1}^{K}
w_t^k \,
p_{\mathcal{M}}(u_t^* \mid h_t^k, \mathcal{H}_t)
\right).
\end{equation}
\paragraph{Score Prediction Loss.}
For each natural-language hypothesis $h_t^k$, we supervise the VLM with an annotated consistency score $r_t^{k*}\in\{1,\ldots,10\}$. The scores are assigned by human annotators and Claude. The model is trained to generate the score as a textual answer using cross-entropy:
\begin{equation}
\mathcal{L}_{\text{score}}
=
-\frac{1}{K}\sum_{k=1}^{K}
\log p_{\mathcal{M}}
\left(r_t^{k*}\mid h_t^k,\boldsymbol{x}_t,\mathcal{H}_{t-1}\right).
\end{equation}
At inference time, the generated scores are rescaled and converted into belief weights using Eqs.~\eqref{eq:weight_unnorm}--\eqref{eq:weight}.
\subsection{Closed-loop Inference and Uncertainty Tracking}
\label{sec:inference}
\begin{figure}[t]
    \centering
    \includegraphics[width=0.95\linewidth]{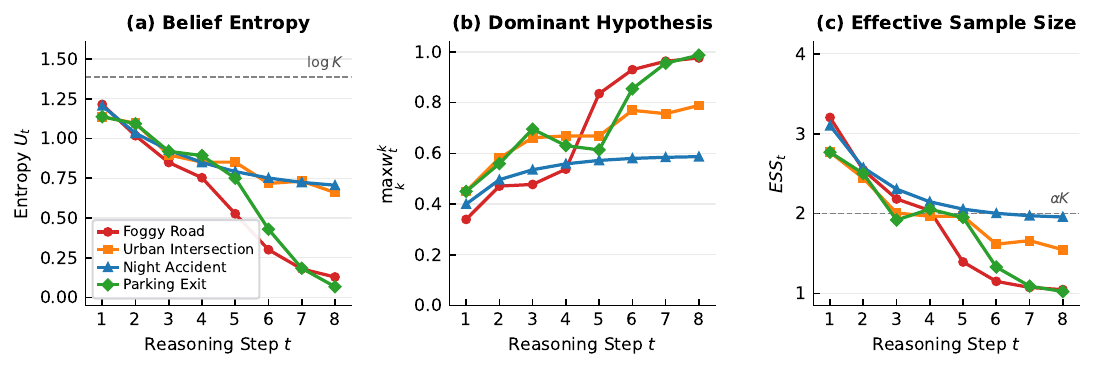}
    \caption{\textbf{Belief dynamics across four scenarios.}
    Foggy Road (red), Urban Intersection (orange), Night Accident (blue), Parking Exit (green).
    \textbf{(a)} Entropy $U_t$ starts near $\log K$ and decays as the scene becomes less ambiguous.
    \textbf{(b)} The dominant weight $\max_k w_t^k$ climbs toward certainty in most scenarios;
    Night Accident plateaus at ${\sim}0.6$, reflecting persistent ambiguity.
    \textbf{(c)} $\text{ESS}_t$ drops below the resampling threshold $\alpha K$ once belief concentrates,
    except in Night Accident where diversity is maintained throughout.
    Per-step visualizations appear in Appendix~\ref{app:qualitative}.}
    \label{fig:belief_dynamics}
\end{figure}

Algorithm~\ref{algo:closed_loop} summarizes the closed-loop inference procedure of DriveMind. We use the entropy of the hypothesis weights to summarize how belief is distributed across hypotheses:
\begin{equation}
\label{eq:entropy}
U_t = -\sum_{k=1}^{K} w_t^k \log(w_t^k).
\end{equation}
Figure~\ref{fig:belief_dynamics} shows how these belief statistics evolve in representative scenarios; additional qualitative examples are in Appendix~\ref{app:qualitative}.
\section{Experiments}
\subsection{Benchmarks}
\label{sec:ben}
We evaluate DriveMind on two autonomous driving benchmarks: Bench2Drive and NAVSIM (Appendix~\ref{app:navsim}). Bench2Drive~\citep{jia2024bench2drive} is a CARLA-based benchmark with 220 short routes, each approximately 150\,m long, spanning Town01--Town15 under varying weather and traffic. We follow the official Bench2Drive evaluation protocol and report standard closed-loop metrics, including Driving Score (DS), Success Rate (SR), efficiency, and comfort. We also report the Bench2Drive multi-ability breakdown, which evaluates performance on merging, overtaking, emergency braking, give-way, and traffic-sign handling, along with the mean score across abilities.
\paragraph{DriveMind w/o Belief.}
This baseline uses the same VLM and diffusion controller as DriveMind but conditions trajectory generation on a single scene interpretation generated from the current observation. It does not propagate hypotheses across reasoning steps or perform belief reweighting and resampling.
\subsection{Results}
\begin{figure*}[t]
    \centering
    \includegraphics[width=\textwidth]{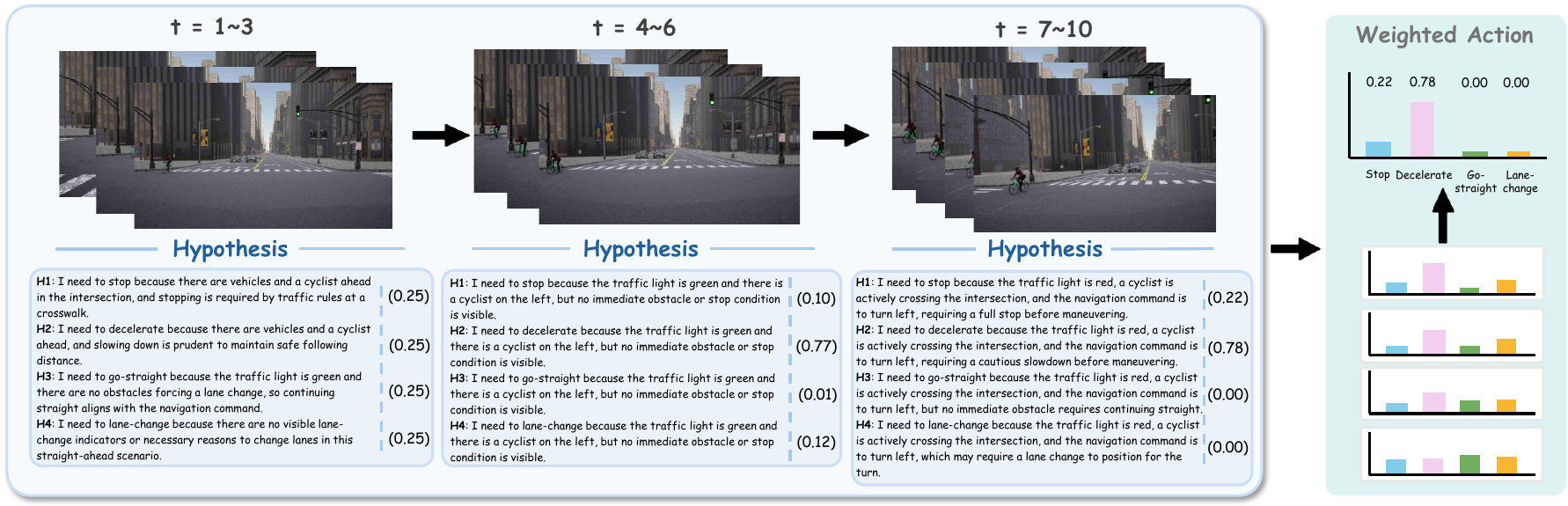}
    \caption{\textbf{Belief evolution and weighted action selection.}
    The ego vehicle approaches an intersection with a left-turn command and a cyclist near the crosswalk. DriveMind starts with four equally weighted hypotheses and concentrates belief on cautious behavior as the cyclist's relevance becomes clear: deceleration receives the highest final weight ($0.78$), with stopping retained as a fallback ($0.22$).}
    \label{fig:case}
\end{figure*}

\paragraph{Driving Performance on Bench2Drive.}
Table~\ref{tab:main} reports the closed-loop driving metrics and multi-ability evaluation results on Bench2Drive.
DriveMind achieves \textbf{81.37} DS and \textbf{54.09\%} SR.
Compared with DriveMind w/o belief, belief tracking improves DS from \textbf{68.50} to \textbf{81.37} (+12.87) and SR from \textbf{41.36\%} to \textbf{54.09\%} (+12.73 percentage points).
Relative to DriveMind w/o belief, the largest multi-ability gains are in give-way (30.00\% to 70.00\%, +40.00 percentage points) and overtaking (57.78\% to 91.11\%, +33.33 percentage points), both of which involve interactions with other road users.
DriveMind also achieves the highest multi-ability mean score among the methods in Table~\ref{tab:main}, at \textbf{58.17\%}.
At the same time, DriveMind does not dominate every individual ability, suggesting that belief tracking is most beneficial in scenarios requiring intent disambiguation rather than rule-following alone.
\begin{table*}[!ht]
\centering
\fontsize{8}{9}\selectfont
\renewcommand{\arraystretch}{0.92}
\caption{Closed-loop evaluation and multi-ability test results on Bench2Drive. Best results are shown in bold.}
\label{tab:main}
\setlength{\tabcolsep}{3pt}
\textbf{(a) Closed-loop metrics $\uparrow$}\\[3pt]
\begin{adjustbox}{max width=\textwidth}
\begin{tabular}{c lcccc}
\toprule
& Method & Efficiency & Comfort & SR (\%) & DS \\
\midrule
\multirow{5}{*}{\rotatebox{90}{w/ dist.}} & TCP~\citep{wu2022trajectory} & 54.26 & 47.80 & 15.00 & 40.70 \\
 & TCP-ctrl~\citep{wu2022trajectory} & 55.97 & \textbf{51.51} & 7.27 & 30.47 \\
 & TCP-traj~\citep{wu2022trajectory} & 76.54 & 18.08 & 30.00 & 59.90 \\
 & ThinkTwice~\citep{jia2023think} & 76.93 & 16.22 & 31.23 & 62.44 \\
 & DriveAdapter~\citep{jia2023driveadapter} & 70.22 & 16.01 & 33.08 & 64.22 \\
\midrule
\multirow{6}{*}{\rotatebox{90}{w/o dist.}} & AD-MLP~\citep{zhai2023rethinking} & 48.45 & 22.63 & 0.00 & 18.05 \\
 & UniAD-T.~\citep{hu2023planning} & 123.92 & 47.04 & 13.18 & 40.73 \\
 & UniAD-B.~\citep{hu2023planning} & 129.21 & 43.58 & 16.36 & 45.81 \\
 & VAD~\citep{jiang2023vad} & \textbf{157.94} & 46.01 & 15.00 & 42.35 \\
 & DriveTransformer-L.~\citep{jia2025drivetransformer} & 100.64 & 46.01 & 35.00 & 63.46 \\
 & ReCogDrive~\citep{li2025recogdrive} & 138.18 & 17.45 & 45.45 & 71.36 \\
\midrule
 & DriveMind w/o belief & $134.75\pm 1.86$ & $18.01\pm 0.08$ & $41.36\pm 1.85$ & $68.50\pm 1.39$ \\
 & \textbf{DriveMind (Ours)} & $114.83\pm 0.33$ & $18.98\pm 0.12$ & $\mathbf{54.09\pm 1.10}$ & $\mathbf{81.37\pm 1.41}$ \\
\bottomrule
\end{tabular}
\end{adjustbox}
\par\medskip
\textbf{(b) Multi-ability test (\%) $\uparrow$}\\[3pt]
\begin{adjustbox}{max width=\textwidth}
\begin{tabular}{c lcccccc}
\toprule
& Method & Merging & Overtaking & Emerg. Brake & Give-way & Traffic Sign & Mean \\
\midrule
\multirow{5}{*}{\rotatebox{90}{w/ dist.}} & TCP & 16.18 & 20.00 & 20.00 & 10.00 & 6.99 & 14.63 \\
 & TCP-ctrl & 10.29 & 4.44 & 10.00 & 10.00 & 6.45 & 8.23 \\
 & TCP-traj & 8.89 & 24.29 & 51.67 & 40.00 & 46.28 & 34.22 \\
 & ThinkTwice & 27.38 & 18.42 & 35.82 & 50.00 & 54.23 & 37.17 \\
 & DriveAdapter & 28.82 & 26.38 & 48.76 & 50.00 & 56.43 & 42.08 \\
\midrule
\multirow{6}{*}{\rotatebox{90}{w/o dist.}} & AD-MLP & 0.00 & 0.00 & 0.00 & 0.00 & 4.35 & 0.87 \\
 & UniAD-T. & 8.89 & 9.33 & 20.00 & 20.00 & 15.43 & 14.73 \\
 & UniAD-B. & 14.10 & 17.78 & 21.67 & 10.00 & 14.21 & 15.55 \\
 & VAD & 8.11 & 24.44 & 18.64 & 20.00 & 19.15 & 18.07 \\
 & DriveTransformer-L. & 17.57 & 35.00 & 48.36 & 40.00 & 52.10 & 38.60 \\
 & ReCogDrive & 29.73 & 20.00 & \textbf{69.09} & 20.00 & \textbf{71.34} & 42.03 \\
\midrule
 & DriveMind w/o belief & $31.25\pm 0.70$ & $57.78\pm 2.77$ & $31.67\pm 1.33$ & $30.00\pm 0.00$ & $34.21\pm 1.47$ & $36.98\pm 1.21$ \\
 & \textbf{DriveMind (Ours)} & $\mathbf{41.25\pm 1.10}$ & $\mathbf{91.11\pm 1.70}$ & $51.67\pm 1.20$ & $\mathbf{70.00\pm 0.00}$ & $36.84\pm 2.01$ & $\mathbf{58.17\pm 1.30}$ \\
\bottomrule
\end{tabular}
\end{adjustbox}
\end{table*}
Figure~\ref{fig:case} shows a representative intersection case where the ego vehicle must decide how to react to a cyclist while following a left-turn command. DriveMind initially keeps multiple hypotheses with equal weights and gradually shifts belief weight toward cautious deceleration as later observations confirm that the cyclist is crossing the intersection.
\paragraph{Scenario Analysis.}
Table~\ref{tab:scenario_category} compares DriveMind with the baseline across nine Bench2Drive scenario categories. The largest DS gains occur in \textit{Opposite Direction} ($+20.5$) and \textit{Unsignalized Intersection} ($+20.0$), where driving decisions depend on uncertain interactions with other agents. \textit{Pedestrian Crossing} ($+16.9$), \textit{Obstacle/Hazard} ($+14.7$), and \textit{Merging} ($+13.6$) also improve substantially, consistent with benefits in interpreting interactions and potential hazards. Gains are smaller in \textit{Emergency/Accident}. SR remains unchanged in \textit{Opposite Direction} ($13.3\%$) and decreases from $12.0\%$ to $4.0\%$ in \textit{Signalized Intersection}.
\paragraph{Long-Tail Scenario Analysis.}
Table~\ref{tab:longtail_bench} examines six challenging scenarios involving pedestrian crossings, merging, and roadside hazards. DriveMind improves DS and SR in all six. The largest gain occurs in \textit{Parking--Crossing Pedestrian}, where DS rises from $53.2$ to $100.0$ and SR from $20\%$ to $100\%$. \textit{Merger into Slow Traffic (V2)} and \textit{Hazard at Side Lane (Two Ways)} gain $35.2$ and $33.4$ DS, respectively. Maintaining competing interpretations may help resolve these ambiguous interactions.
\begin{table}[t]
\centering
\footnotesize
\caption{Results across nine Bench2Drive scenario categories. $\Delta$DS is the gain over DriveMind w/o belief.}
\label{tab:scenario_category}
\setlength{\tabcolsep}{4pt}
\begin{tabular}{lcccccc}
\toprule
\multirow{2}{*}{Scenario Category}
& \multicolumn{2}{c}{DS $\uparrow$}
& \multicolumn{2}{c}{SR (\%) $\uparrow$}
& \multirow{2}{*}{$\Delta$DS} \\
\cmidrule(lr){2-3}\cmidrule(lr){4-5}
& DriveMind w/o belief & DriveMind
& DriveMind w/o belief & DriveMind
& \\
\midrule
Unsignalized Intersection   & 54.4 & 74.4 & 16.7 & 30.0 & \textbf{+20.0} \\
Opposite Direction          & 42.7 & 63.2 & 13.3 & 13.3 & \textbf{+20.5} \\
Pedestrian Crossing         & 53.3 & 70.2 & 20.0 & 30.0 & +16.9 \\
Obstacle / Hazard           & 84.0 & 98.7 & 73.3 & 96.7 & +14.7 \\
Merging                     & 72.0 & 85.6 & 50.0 & 70.0 & +13.6 \\
Cut-in / Lane Change        & 81.6 & 94.0 & 60.0 & 85.0 & +12.4 \\
Actor Flow / Highway        & 73.6 & 80.8 & 40.0 & 55.0 & +7.2  \\
Signalized Intersection     & 58.8 & 64.6 & 12.0 &  4.0 & +5.8  \\
Emergency / Accident        & 88.0 & 91.7 & 76.0 & 84.0 & +3.7  \\
\midrule
\textbf{Overall}            & 68.5 & 81.4 & 41.4 & 54.1 & \textbf{+12.9} \\
\bottomrule
\end{tabular}
\end{table}
\begin{table}[t]
\centering
\small
\caption{Results on six selected challenging Bench2Drive scenarios. Baseline denotes DriveMind w/o belief; SR is reported in percent.}
\label{tab:longtail_bench}
\setlength{\tabcolsep}{4pt}
\begin{tabular}{lcccccc}
\toprule
\multirow{2}{*}{Scenario} & \multicolumn{2}{c}{DS $\uparrow$} & \multicolumn{2}{c}{SR (\%) $\uparrow$} & \multirow{2}{*}{$\Delta$DS} \\
\cmidrule(lr){2-3}\cmidrule(lr){4-5}
& Baseline & DriveMind & Baseline & DriveMind & \\
\midrule
Parking--Crossing Pedestrian         &  53.2 & 100.0 &  20 & 100 & \textbf{+46.8} \\
Merger into Slow Traffic (V2)        &  44.0 &  79.2 &   0 &  60 & \textbf{+35.2} \\
Hazard at Side Lane (Two Ways)       &  66.6 & 100.0 &  40 & 100 & \textbf{+33.4} \\
Non-Signalized Junction Right Turn   &  49.7 &  81.6 &  20 &  40 & +31.9 \\
T-Junction                           &  59.2 &  86.0 &   0 &  60 & +26.8 \\
Interurban Advanced Actor Flow       &  68.0 &  92.0 &  20 &  80 & +24.0 \\
\bottomrule
\end{tabular}
\end{table}
\subsection{Ablation Studies}
\label{sec:ablation}
\begin{figure*}[!t]
    \centering
    \includegraphics[width=\textwidth]{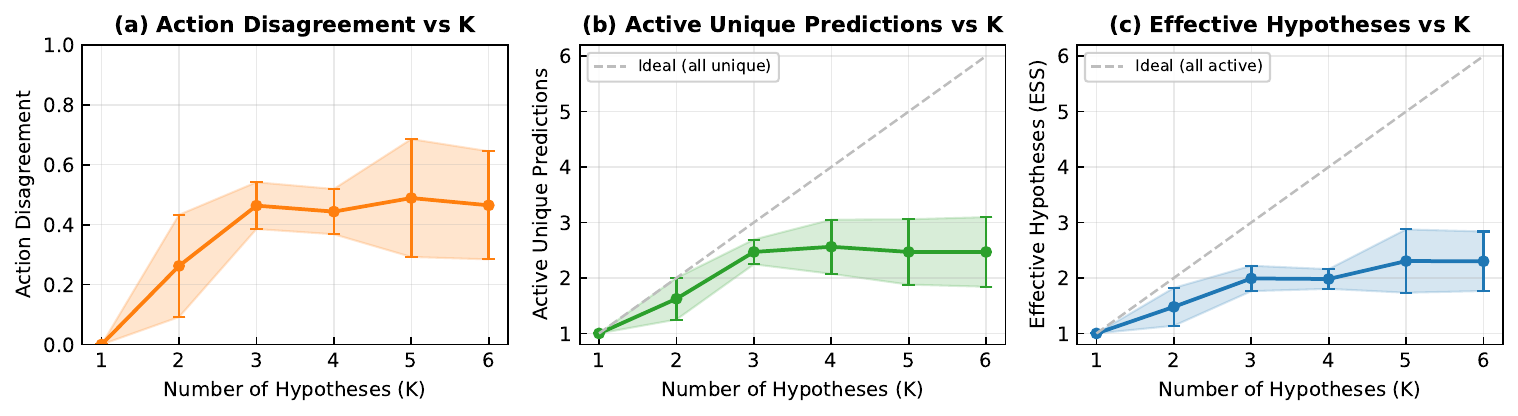}
    \caption{\textbf{Effect of hypothesis count $K$ on behavioral diversity.}
    (a) Weighted action disagreement. (b) Maneuver categories with total weight at least $0.05$. (c) ESS after reweighting; dashed lines mark $K$. Curves show means across four scenarios, each averaged over eight reasoning steps; shaded bands and error bars show one standard deviation across scenario means.}
    \label{fig:k_behavior}
\end{figure*}

We conduct ablation studies to evaluate four key design choices in DriveMind: multi-hypothesis belief tracking, the hypothesis count, scoring mode and inference latency.
\paragraph{Effect of Multi-Hypothesis Belief Tracking.}
Table~\ref{tab:ablation_component} compares the single-interpretation baseline with multi-hypothesis belief tracking. With $K{=}2$, DS increases from $68.50$ to $70.00$ and SR from $41.36\%$ to $48.00\%$; with $K{=}4$, they reach $81.37$ and $54.09\%$, respectively. The $K{=}2$ setting initializes hypotheses for going straight and stopping, while $K{=}4$ additionally includes deceleration and lane change. Thus, this comparison varies both hypothesis count and initial maneuver coverage.
\begin{table}[t]
\centering
\small
\begin{minipage}[t]{0.48\linewidth}
\centering
\footnotesize
\captionof{table}{Ablation of multi-hypothesis belief tracking on Bench2Drive.}
\label{tab:ablation_component}
\setlength{\tabcolsep}{5pt}
\begin{tabular}{lcc}
\toprule
Method & DS $\uparrow$ & SR (\%) $\uparrow$ \\
\midrule
DriveMind w/o belief & 68.50 & 41.36 \\
DriveMind ($K{=}2$)      & 70.00 & 48.00 \\
DriveMind ($K{=}4$)      & \textbf{81.37} & \textbf{54.09} \\
\bottomrule
\end{tabular}
\end{minipage}
\hfill
\begin{minipage}[t]{0.48\linewidth}
\centering
\footnotesize
\captionof{table}{Effect of weight source on maneuver prediction accuracy. }
\label{tab:score_mode}
\setlength{\tabcolsep}{7pt}
\begin{tabular}{lcc}
\toprule
Mode & Maneuver Acc. $\uparrow$ & Entropy \\
\midrule
Random & 0.644 & 1.090 \\
Uniform ($1/K$) & 0.779 & 1.386 \\
VLM score (ours) & \textbf{0.930} & \textbf{0.711} \\
\bottomrule
\end{tabular}
\end{minipage}
\end{table}
\begin{table}[t]
\centering
\footnotesize
\caption{
Runtime analysis of DriveMind on a single NVIDIA RTX A6000 GPU.
Left: isolated component latency. Right: system-level latency under shared-GPU execution. P95 denotes the 95th-percentile latency. $s$ denotes the number of fast control steps between two slow reasoner updates.
}
\label{tab:runtime_async}
\setlength{\tabcolsep}{4pt}
\resizebox{0.95\linewidth}{!}{
\begin{tabular}{lcc@{\hspace{1.0em}}lcc}
\toprule
\multicolumn{3}{c}{\textit{Component-level latency}} &
\multicolumn{2}{c}{\textit{System-level latency}} \\
\cmidrule(r){1-3} \cmidrule(l){4-5}
Component & Mean (ms) & P95 (ms) &
Setting & Mean (ms) \\
\midrule
High Level Hypothesis Evolving ($K{=}2$) & 74.0 & 77.9 &
Synchronous baseline & 177.0  \\
High Level Hypothesis Evolving ($K{=}4$) & 120.0 & 125.4 &
Async refresh ($s{=}2$) & 94.7  \\
High Level Hypothesis Evolving ($K{=}6$) & 165.5 & 167.6 &
Async refresh ($s{=}4$) & 65.5  \\
Low-level control & 45.0 & 45.0 &
Async refresh ($s{=}6$) & 57.1  \\
\bottomrule
\end{tabular}
}
\end{table}
\paragraph{Effect of Hypothesis Count.}
We further analyze how the number of maintained hypotheses $K$ affects behavioral diversity during belief tracking. Figure~\ref{fig:k_behavior} reports three statistics as $K$ varies. Let $u_t^k$ be the maneuver label of hypothesis $h_t^k$, and define its aggregated action weight as $\pi_t(u)=\sum_{k=1}^{K}w_t^k\mathbf{1}[u_t^k=u]$. We measure (i) \textbf{action disagreement} as $1-\sum_{u\in\mathcal{U}}\pi_t(u)^2$; (ii) \textbf{active unique predictions} as the number of maneuver categories with $\pi_t(u)\geq0.05$; and (iii) \textbf{effective hypotheses} using $ESS_t$. As $K$ increases, action disagreement and the number of unique predictions grow rapidly until $K{\approx}4$, after which they plateau. Similarly, the effective number of hypotheses grows sublinearly with $K$, indicating that the weights are not uniformly distributed across hypotheses. The plateau in unique maneuver predictions should be interpreted in light of the fixed four-category vocabulary: $K{=}4$ already covers all maneuver categories at initialization. This saturation is therefore consistent with the chosen maneuver partition, rather than evidence of unrestricted semantic coverage. We use $K{=}4$ in all main experiments.
\paragraph{Effect of Scoring Mode.}
We evaluate maneuver prediction accuracy using the most probable maneuver under the belief-weighted marginal of Eq.~\eqref{eq:maneuver_marginal}:
\begin{equation}
\hat{u}_t = \arg\max_{u\in\mathcal{U}} \sum_{k=1}^{K} w_t^k p_{\mathcal{M}}(u\mid h_t^k,\mathcal{H}_t),
\qquad
\mathrm{Acc.} = \frac{1}{|\mathcal{T}_{\mathrm{eval}}|}\sum_{t\in\mathcal{T}_{\mathrm{eval}}}\mathbf{1}[\hat{u}_t=u_t^*],
\label{eq:dominant_accuracy}
\end{equation}
where $\mathcal{T}_{\mathrm{eval}}$ is the set of evaluated frames and $u_t^*$ is the reference maneuver derived from the expert autopilot's recorded state and controls, using the four-way vocabulary in Appendix~\ref{app:maneuver}.
We compare VLM-based weighting with uniform weights ($w_t^k=1/K$) and random weights, keeping the hypothesis set and reference labels fixed. As shown in Table~\ref{tab:score_mode}, VLM-based weighting achieves $93.0\%$ maneuver prediction accuracy, compared with $77.9\%$ for uniform weighting and $64.4\%$ for random weighting. These results support the usefulness of consistency-based reweighting. The lower entropy ($0.711$ vs.\ $1.386$ for uniform weighting) indicates more concentrated weights, rather than calibrated confidence.
\paragraph{Inference Latency.}
\label{sec:latency}
Table~\ref{tab:runtime_async} reports both isolated component latency and system-level latency on a single NVIDIA RTX A6000 GPU. The high-level belief update remains within the slow-reasoner budget even as the hypothesis count increases: it takes $74.0$ ms for $K{=}2$, $120.0$ ms for the default $K{=}4$, and $165.5$ ms for $K{=}6$. The low-level controller takes $45.0$ ms, below the $100$ ms budget of a 10 Hz control loop. We also profile shared-GPU execution. A synchronous baseline takes $177.0$ ms per step, while asynchronous belief refresh reduces the average control-step latency to $94.7$ ms at $s{=}2$, $65.5$ ms at $s{=}4$, and $57.1$ ms at $s{=}6$. These results show that the reasoning module can be updated at low frequency without dominating runtime, and that asynchronous refresh substantially reduces control-side latency compared with blocking reasoning updates.
\section{Conclusion and Limitations}
\label{sec:conclusion}
We presented \textit{DriveMind}, a closed-loop MLLM-based driving framework that maintains a time-evolving belief state over multiple natural-language hypotheses. By propagating and reweighting hypotheses as new observations arrive, DriveMind reduces premature commitment under ambiguity. Its asynchronous fast--slow architecture decouples slow VLM-based belief tracking from fast diffusion-based trajectory generation. Experiments on Bench2Drive confirm improved performance and robustness in ambiguous scenarios. Ablations support the effectiveness of multi-hypothesis belief tracking, while runtime analysis demonstrates the efficiency of asynchronous inference.
Despite these promising results, several directions remain open. The hypothesis count $K$ and update frequency are currently fixed; adapting them based on real-time scene complexity or belief entropy could improve efficiency and responsiveness. Incorporating predictions over other agents' future intentions as part of the hypothesis space, rather than focusing on ego-vehicle maneuvers alone, would allow the belief state to reason more explicitly about multi-agent interactions.
\section*{Acknowledgments}
This work was supported in part by the Pacific Northwest Transportation Consortium (PacTrans), funded by the U.S. Department of Transportation (USDOT). We thank Dr. Ryan Feng Lin for his insightful discussions and valuable feedback.
{
\small
\bibliographystyle{plainnat}
\bibliography{neurips}
}
\newpage
\appendix
\section{Implementation Details}
\subsection{Model Architecture}
DriveMind consists of a high-level VLM reasoner and a low-level diffusion controller.
For the high-level reasoner, we use Qwen3VL-2B-Instruct~\citep{bai2025qwen3vl} as the backbone.
The visual encoder adopts a ViT-L/14 architecture with an input resolution of $448 \times 448$.
The VLM generates natural-language hypotheses, predicts maneuver distributions, and produces belief scores for hypothesis reweighting.
For the low-level controller, we use a DDPM-based diffusion policy~\citep{ho2020denoising} with a Transformer-based backbone.
The controller is conditioned on the belief context $c_t$ and visual feature history $\Phi(\mathcal{H}_t)$ to generate short-horizon waypoint trajectories.
\subsection{Training the Low-Level Diffusion Controller}
\label{app:diff_train}
We train the diffusion controller $\mathcal{D}$ to generate a continuous future waypoint trajectory $a_t$ conditioned on the visual feature history $\Phi(\mathcal{H}_t)$ and the belief context $c_t$.
\paragraph{Decoupled Training Strategy.}
After training the high-level reasoner, we freeze $\mathcal{M}$ and pre-compute belief contexts for controller learning.
\textbf{Phase I: Belief Caching (Offline).}
Using the frozen $\mathcal{M}$, we run belief tracking over the training sequences and compute $c_t$ at each timestep.
We cache $(c_t, \Phi(\mathcal{H}_t), a_t^{0})$, where $a_t^{0}$ denotes the expert waypoint trajectory.
\textbf{Phase II: Controller Training.}
We train $\mathcal{D}$ by minimizing the conditional diffusion denoising objective:
\begin{equation}
\mathcal{L}_{\text{diff}}
=
\mathbb{E}_{\tau, \epsilon, a_t^{0}}
\left[
\left\|
\epsilon -
\epsilon_\theta(a_t^\tau, \tau, c_t, \Phi(\mathcal{H}_t))
\right\|_2^2
\right].
\label{eq:diff_loss}
\end{equation}
where $a_t^\tau$ is obtained by adding Gaussian noise to the expert trajectory $a_t^{0}$ at diffusion step $\tau$.
During inference, the two subsystems are coupled online: $\mathcal{M}$ updates the belief state $\mathcal{B}_t$ and predicts the maneuver distribution, while $\mathcal{D}$ consumes the latest belief context $c_t$ to synthesize a trajectory in real time.
\subsection{Training Details}
DriveMind is trained in two stages.
In the first stage, we fine-tune the VLM using a mixture of VQA and action-prediction samples with a 1:1 sampling ratio.
The VLM is optimized with maneuver (STOP, DECELERATE, GO\_STRAIGHT, LANE\_CHANGE) prediction and score prediction losses. For score prediction, we apply cross-entropy to the annotated score answers for individual natural-language hypotheses.
In the second stage, we freeze the VLM and pre-compute belief contexts over the training sequences.
The diffusion controller is then trained to denoise expert waypoint trajectories conditioned on the cached belief context and visual features.
All experiments are conducted on 2 NVIDIA A6000 GPUs.
The training hyperparameters are summarized in Table~\ref{tab:hyper}.
\begin{table}[h]
\centering
\caption{Training hyperparameters for DriveMind.}
\label{tab:hyper}
\begin{tabular}{lc|lc}
\toprule
\textbf{Parameter} & \textbf{Value} & \textbf{Parameter} & \textbf{Value} \\
\midrule
VLM Learning Rate & $2\times10^{-5}$ & Diffusion Learning Rate & $1\times10^{-4}$ \\
VLM Batch Size & 8 & Diffusion Batch Size & 256 \\
Weight Decay (VLM) & 0.0001 & DDPM Steps & 20 \\
Max Sequence Length & 2048 & Weight Decay (Diffusion) & 0.0005 \\
\bottomrule
\end{tabular}
\end{table}
\subsection{Closed-loop Inference Algorithm}
\label{app:algorithm}
\vspace{-0.5em}
\begin{algorithm}[H]
\small
\caption{Closed-loop Belief Tracking and Control (Inference)}
\label{algo:closed_loop}
\begin{algorithmic}[1]
\Require Reasoning model $\mathcal{M}$, diffusion controller $\mathcal{D}$, observations $\boldsymbol{x}_{1:T}$, hypothesis count $K$, ESS threshold $\alpha$, similarity threshold $\rho$, temperature $\beta$.
\State Initialize trace $\tau=\emptyset$, hypotheses $\{h_0^k\}_{k=1}^{K}$, weights $w_0^k=1/K$, and belief state $\mathcal{B}_0=\{(h_0^k,w_0^k)\}_{k=1}^{K}$.
\For{$t=1,\dots,T$}
    \State Construct history $\mathcal{H}_{t-1}=(\boldsymbol{x}_{1:t-1},a_{1:t-2})$.
    \Statex \Comment{\textit{--- Predict: Hypothesis Propagation (Slow) ---}}
    \For{$k=1,\dots,K$}
        \State Sample $h_t^k \sim p_{\mathcal{M}}(\cdot \mid h_{t-1}^k,a_{t-1},\boldsymbol{x}_t,\mathcal{H}_{t-1})$. \Comment{Eq.~\eqref{eq:propagate}}
    \EndFor
    \Statex \Comment{\textit{--- Correct: Evidence-Consistency Reweighting (Slow) ---}}
    \For{$k=1,\dots,K$}
        \State Compute score $s_t^k=f_{\mathcal{M}}(h_t^k,\boldsymbol{x}_t,\mathcal{H}_{t-1})$. \Comment{Eq.~\eqref{eq:score}}
        \State Compute unnormalized weight $\tilde{w}_t^k=w_{t-1}^k\exp(s_t^k/\beta)$. \Comment{Eq.~\eqref{eq:weight_unnorm}}
    \EndFor
    \State Normalize weights $w_t^k=\tilde{w}_t^k / \sum_{j=1}^{K}\tilde{w}_t^j$. \Comment{Eq.~\eqref{eq:weight}}
    \Statex \Comment{\textit{--- Resample: Diversity Maintenance (Slow) ---}}
    \State Compute $ESS_t$ via Eq.~\eqref{eq:ESS}.
    \If{$ESS_t < \alpha K$}
        \State Resample $K$ hypotheses from $\{(h_t^k,w_t^k)\}$ according to $\{w_t^k\}$.
        \State Compute pairwise cosine similarity $\mathrm{sim}(j,k)$ using normalized VLM text embeddings.
        \For{each pair $(j,k)$ with $\mathrm{sim}(j,k)>\rho$}
            \State Keep the higher-weighted hypothesis and regenerate the other so that its similarity to the kept hypothesis is at most $\rho$.
        \EndFor
        \State Reset all weights to $w_t^k=1/K$ and update $\mathcal{B}_t=\{(h_t^k,w_t^k)\}_{k=1}^{K}$.
    \Else
        \State Set $\mathcal{B}_t=\{(h_t^k,w_t^k)\}_{k=1}^{K}$.
    \EndIf
    \Statex \Comment{\textit{--- Control: Belief-Conditioned Trajectory Generation (Fast) ---}}
    \State Construct current history $\mathcal{H}_t=(\boldsymbol{x}_{1:t},a_{1:t-1})$.
    \State Compute belief context $c_t=\sum_{k=1}^{K}w_t^k E_{\mathcal{M}}(h_t^k)$. \Comment{Eq.~\eqref{eq:belief_context}}
    \State Sample trajectory $a_t\sim p_{\mathcal{D}}(a\mid c_t,\Phi(\mathcal{H}_t))$ via reverse diffusion.
    \State Execute $a_t$ in a receding-horizon manner.
    \State Compute uncertainty $U_t=-\sum_{k=1}^{K}w_t^k\log w_t^k$.
    \State Add $(\mathcal{B}_t,c_t,a_t,U_t,ESS_t)$ to trace $\tau$.
\EndFor
\State \Return $\tau$.
\end{algorithmic}
\end{algorithm}
\subsection{Diversity Maintenance Visualization}
\label{app:diversity}
\begin{figure}[t]
    \centering
    \includegraphics[width=0.9\linewidth]{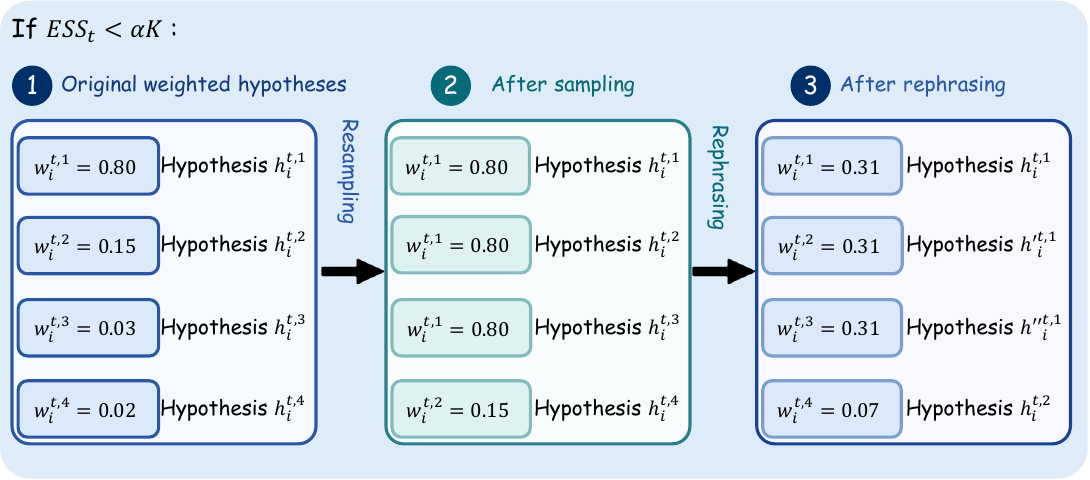}
    \caption{\textbf{Diversity maintenance through resampling and regeneration.}
    When the effective sample size $ESS_t$ falls below the threshold $\alpha K$, DriveMind resamples high-weight hypotheses and regenerates near-duplicated hypotheses to maintain semantic diversity while preserving consistency with the current observation.}
    \label{fig:diversity_maintenance}
\end{figure}

Figure~\ref{fig:diversity_maintenance} illustrates the diversity-maintenance step used after evidence-consistency reweighting. When the belief distribution becomes highly concentrated, naive resampling may repeatedly select the same high-weight hypothesis, reducing the diversity of the belief state. DriveMind addresses this issue by regenerating near-duplicated hypotheses into semantically distinct alternatives while keeping them consistent with the current observation.
\section{Formal Definitions of Evaluation Metrics}
\label{app:metrics}
To provide a rigorous evaluation of \textit{DriveMind}, we employ two sets of metrics corresponding to the Bench2Drive and NAVSIM benchmarks. These metrics evaluate the agent's performance from the perspectives of safety, efficiency, comfort, and similarity to human driving behavior.
\subsection{Bench2Drive (CARLA-based) Metrics}
The closed-loop performance in the CARLA simulator is primarily characterized by the Driving Score (DS) and Success Rate (SR).
\paragraph{Driving Score (DS):}
The DS is the primary metric in Bench2Drive, defined as the product of the Route Completion (RC) and the Infraction Penalty (IP):
\begin{equation}
DS = \frac{1}{N} \sum_{i=1}^{N} RC_i \times IP_i,
\end{equation}
where $RC_i$ is the percentage of the $i$-th route completed by the agent. $IP_i$ is a cumulative penalty factor that starts at 1.0 and is reduced whenever an infraction occurs (e.g., collisions with pedestrians, vehicles, or static objects, and red-light violations).
\paragraph{Success Rate (SR):}
SR is the ratio of routes completed without any infractions:
\begin{equation}
SR = \frac{\text{Number of Successful Routes}}{\text{Total Number of Routes}}.
\end{equation}
\paragraph{Infraction Penalty (IP):}
Each type of infraction has a specific weight $\omega_{inf}$. If multiple infractions occur, their penalties are multiplied: $IP = \prod \omega_{inf}$. Common penalty weights include:
\begin{itemize}
    \item \textbf{Collision (Pedestrian/Vehicle):} 0.50
    \item \textbf{Collision (Static Layout):} 0.65
    \item \textbf{Red Light Violation:} 0.70
\end{itemize}
\subsection{NAVSIM (PDM-based) Metrics}
 For NAVSIM~\citep{dauner2024navsim}, we follow the official Predictive Driver Model Score (PDMS), which summarizes multiple aspects of closed-loop planning quality, including collision avoidance, drivable-area compliance, time-to-collision safety, comfort, and route progress. The score is computed as
\begin{equation}
\mathrm{PDMS}
=
\mathrm{NC}
\cdot
\mathrm{DAC}
\cdot
\left(
\frac{
5\,\mathrm{EP}
+
5\,\mathrm{TTC}
+
2\,\mathrm{C}
}{12}
\right).
\end{equation}
Where NC denotes the no-collision score, DAC denotes drivable-area compliance, EP denotes ego progress, TTC denotes time-to-collision safety, and C denotes comfort.
\section{NAVSIM Results}
\label{app:navsim}
We further evaluate DriveMind on NavSim~\citep{dauner2024navsim} to test whether the planner generalizes beyond CARLA. Table~\ref{tab:navtest_exp} reports the NavSim sub-scores and the overall PDMS.
DriveMind reaches $81.0$ PDMS, slightly below TransFuser ($84.0$) and UniAD ($83.4$). The sub-scores show that DriveMind remains competitive on safety-related metrics: it achieves $96.5$ NC, $91.1$ TTC, and $100.0$ comfort. The remaining gap mainly comes from DAC and EP, where DriveMind is slightly weaker than the strongest single-step planners.
Importantly, the ablation in Table~\ref{tab:navtest_exp} shows that this gap is not caused by the belief module. Removing belief tracking gives $78.2$ PDMS, while adding belief tracking improves the score to $81.0$. The improvement is small but consistent: DAC increases from $88.9$ to $90.6$, EP increases from $75.0$ to $76.1$, and the safety-related metrics remain unchanged.
This observation is consistent with the different roles of the two benchmarks. NavSim is a non-reactive, pseudo closed-loop benchmark: the planner is evaluated against logged sensor data, and the environment does not change in response to the planned trajectory. As a result, the main source of performance is the quality of the current predicted trajectory, while the benefit of propagating and reweighting multiple hypotheses across future interactive frames is naturally limited. In contrast, Bench2Drive evaluates the agent in an interactive simulator, where early misinterpretations can affect future observations and require recovery. Under this setting, belief tracking provides much larger gains, especially in ambiguous and interaction-heavy scenarios.
We therefore interpret the NavSim result as complementary to the Bench2Drive result. On NavSim, belief tracking provides a small positive gain and does not harm single-step planning performance. On Bench2Drive, where closed-loop interaction makes premature commitment more costly, the same belief mechanism yields substantially larger improvements.
\begin{table}[t]
  \centering
  \caption{Driving performance on Navsim}
  \label{tab:navtest_exp}
  \begin{tabular}{lcccccc}
    \toprule
    Method & NC$\uparrow$ & DAC$\uparrow$ & TTC$\uparrow$ & Comf.$\uparrow$ & EP$\uparrow$ & PDMS$\uparrow$ \\
    \midrule
    Constant Velocity & 68.0 & 57.8 & 50.0 & 100 & 19.4 & 20.6 \\
    Ego Status MLP & 93.0 & 77.3 & 83.6 & 100 & 62.8 & 65.6 \\
    VADv2-$V_{8192}$~\citep{jiang2026vadv2} & 97.2 & 89.1 & 91.6 & 100 & 76.0 & 80.9 \\
    DrivingGPT~\citep{chen2025drivinggpt} & \textbf{98.9} & 90.7 & \textbf{94.9} & 95.6 & \textbf{79.7} & 82.4 \\
    UniAD~\citep{hu2023planning} & 97.8 & 91.9 & 92.9 & 100 & 78.8 & 83.4 \\
    TransFuser~\citep{chitta2022transfuser} & 97.7 & \textbf{92.8} & 92.8 & 100 & 79.2 & \textbf{84.0} \\
    \midrule
    DriveMind w/o belief & 94.3 & 88.9 & 86.7 & 100.0 & 75.0 & 78.2 \\
    DriveMind w belief & 96.5 & 90.6 & 91.1 & 100.0 & 76.1 & 81.0 \\
    \bottomrule
  \end{tabular}
\end{table}
\section{Prompt and Maneuver Space}
\label{app:prompt_maneuver}
\paragraph{Hypothesis propagation prompt.}
\label{app:prompt1}
For each particle $k$, the VLM receives the previous hypothesis $h_{t-1}^k$, the previous executed action $a_{t-1}$, and the current camera observation $\boldsymbol{x}_t$, and generates one updated belief $h_t^k$.
The propagation prompt is:
\begin{promptbox}
You are the high-level reasoning module of an autonomous driving system.\\
Your task is to maintain one plausible belief about the current traffic scene and the immediate driving intention of the ego vehicle.\\[4pt]
Previous belief about the scene: \{$h_{t-1}^k$\}\\
Previous action taken by the ego vehicle: \{$a_{t-1}$\}\\
Current speed: \{speed\} m/s\\
History trajectory: \{trajectory history\}\\[4pt]
Based on the current camera image, update this belief.\\
Focus on:\\
-- the most relevant nearby agents or obstacles,\\
-- whether they are yielding, crossing, merging, braking, or turning,\\
-- what the ego vehicle should do next at a high level.\\[4pt]
Output exactly one hypothesis in 1--2 sentences, then output one maneuver label from \{STOP, DECELERATE, GO\_STRAIGHT, LANE\_CHANGE\} on a new line.
\end{promptbox}
\paragraph{Evidence-consistency scoring prompt.}
\label{app:prompt2}
After propagation, each updated hypothesis $h_t^k$ is evaluated by a second VLM pass that checks whether it is consistent with the current observation and a reference action.
The scoring prompt is:
\begin{promptbox}
You are verifying whether a driving belief is consistent with the current scene.\\[4pt]
Belief: \{$h_t^k$\}\\
Given the current camera image, rate how consistent this belief is with what the ego vehicle should do now.
Answer with exactly one integer from 1 to 10.
\end{promptbox}
We rescale the integer output to $[0,1]$ and use it as the scalar evidence-consistency score $s_t^k$.
\paragraph{Maneuver space.}
\label{app:maneuver}
We use a four-way maneuver vocabulary:
\begin{equation}
    \mathcal{U} = \{\texttt{STOP},\, \texttt{DECELERATE},\, \texttt{GO\_STRAIGHT},\, \texttt{LANE\_CHANGE}\}.
\end{equation}
\texttt{STOP} triggers an emergency brake, \texttt{DECELERATE} covers cautious slowing for hazards or yielding, \texttt{GO\_STRAIGHT} denotes unimpeded forward progress, and \texttt{LANE\_CHANGE} captures lateral maneuvers including overtaking and branch selection at intersections.
\paragraph{Reference Maneuver Labels.}
For the accuracy metric in Eq.~\eqref{eq:dominant_accuracy}, reference labels are constructed from expert autopilot recordings using ordered rules. LANE\_CHANGE is assigned first using lateral navigation commands or steering. Among the remaining frames, STOP is identified using low speed and braking; DECELERATE is assigned using braking while moving or an active \texttt{should\_brake} flag. Remaining frames are labeled GO\_STRAIGHT.
\section{Additional Qualitative Results and Failure Cases}
\label{app:qualitative}
\begin{figure*}[htbp]
    \centering
    \includegraphics[width=\textwidth]{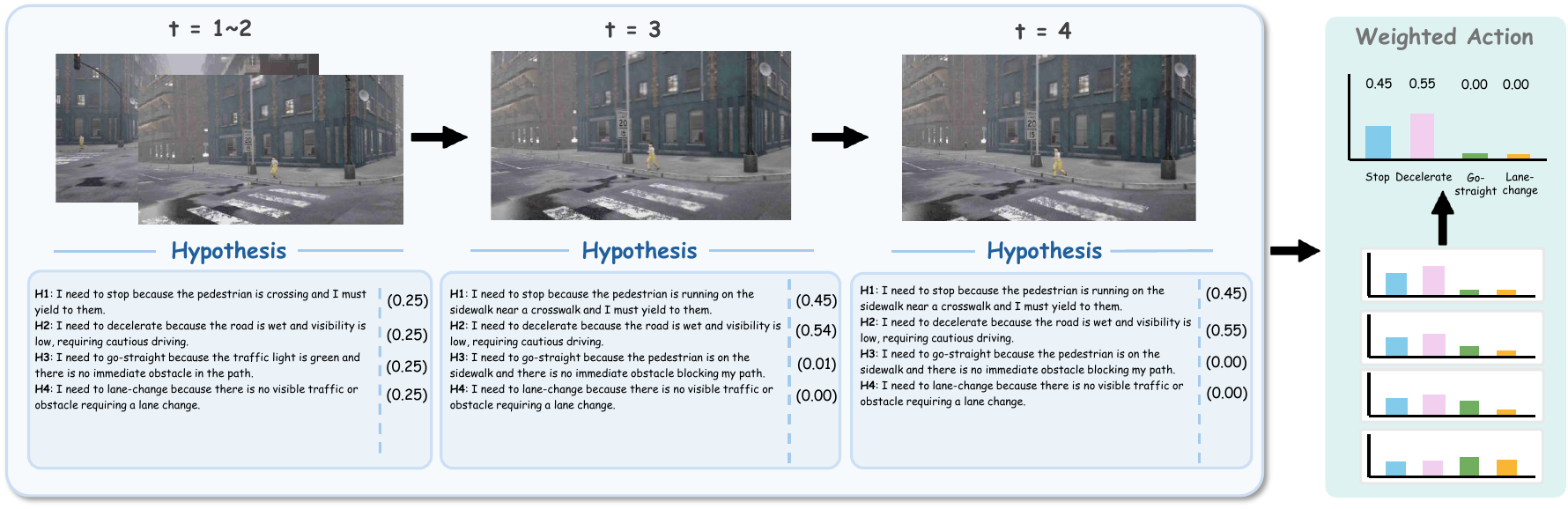}
    \caption{\textbf{Intersection uncertainty.}
    DriveMind initially maintains multiple plausible actions under ambiguity.
     As the pedestrian remains near the crosswalk and wet-road conditions persist, the belief concentrates on cautious behavior, assigning final weights of $0.55$ to deceleration and $0.45$ to stopping, while go-straight and lane-change actions are down-weighted to nearly zero.}
    \label{fig:case_left}
\end{figure*}

\begin{figure*}[htbp]
    \centering
    \includegraphics[width=\textwidth]{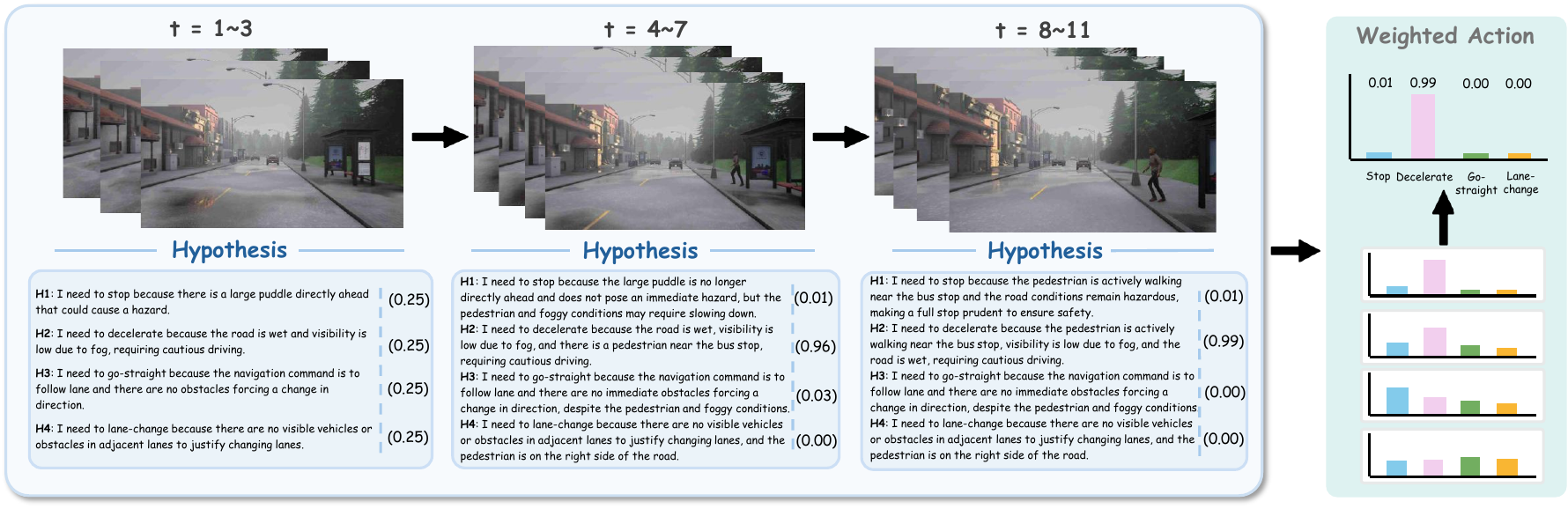}
    \caption{\textbf{Foggy weather with pedestrian uncertainty.}
    DriveMind handles a foggy and wet-road scenario where a pedestrian appears near the roadside.
    Early hypotheses include stopping for a potential hazard, decelerating due to low visibility, continuing straight, and lane changing.
    As the scene evolves, the belief concentrates on the deceleration hypothesis, reaching weight $0.99$ at the final timestep.
    This illustrates how belief tracking supports cautious control under degraded visibility and pedestrian uncertainty.}
    \label{fig:case_fog_pedestrian}
\end{figure*}

\begin{figure}[t]
    \centering
    \includegraphics[width=\columnwidth]{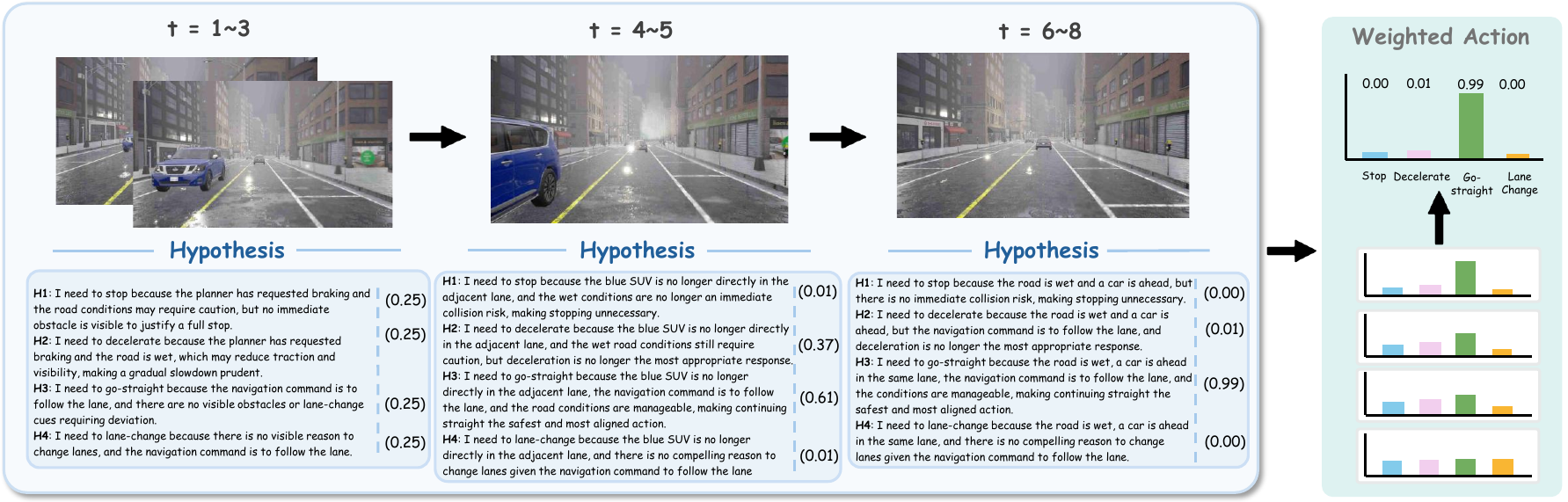}
    \caption{\textbf{Stable parking exit behavior in a low-ambiguity scenario.}
    DriveMind initially maintains multiple plausible hypotheses because of wet road conditions and a nearby vehicle.
As the front vehicle moves farther ahead and the lane remains clear, DriveMind quickly concentrates its belief on the go-straight hypothesis.
At the final timestep, the weighted action assigns $0.99$ to go-straight, while stopping, deceleration, and lane-change are down-weighted to nearly zero.
This case shows that DriveMind does not remain overly cautious when the visual evidence supports safe forward progress.}
    \label{fig:case_carfollow}
\end{figure}

\begin{figure}[t]
    \centering
    \includegraphics[width=\columnwidth]{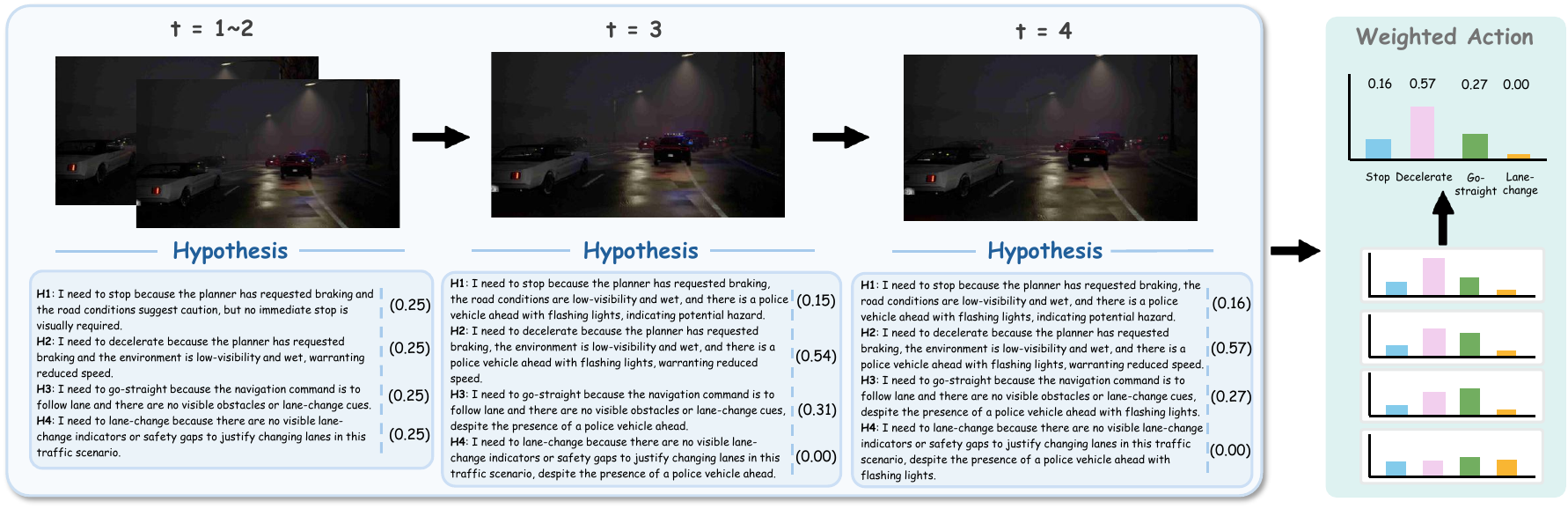}
    \caption{\textbf{Nighttime driving with accident uncertainty.}
    The ego vehicle drives in a dark, wet, and low-visibility scene with a police vehicle ahead.
    DriveMind initially maintains multiple plausible hypotheses with nearly uniform weights, including stopping, decelerating, going straight, and lane changing.
    As new observations confirm persistent poor visibility and the presence of the emergency vehicle, the belief shifts toward conservative behavior, with the final weighted action assigning the highest probability to deceleration while still retaining smaller probabilities for stopping and going straight.}
    \label{fig:case_night}
\end{figure}

\makeatletter
\setlength{\@fptop}{0pt}
\setlength{\@fpsep}{24pt}
\makeatother
We provide additional qualitative case studies to illustrate how DriveMind updates its hypotheses and belief weights over time.
Each case visualizes the observed scene sequence, the maintained natural-language hypotheses, their belief weights, and the resulting weighted action distribution. Figures~\ref{fig:case_left}, \ref{fig:case_fog_pedestrian}, \ref{fig:case_carfollow}, and~\ref{fig:case_night} present representative cases covering intersection uncertainty, foggy-weather pedestrian uncertainty, nighttime accident uncertainty, and stable parking exit behavior.
\paragraph{Failure Cases.}
Despite strong overall performance, DriveMind can still fail in highly challenging scenarios, including extremely dense multi-agent interactions, sudden appearance of fast-moving objects, and prolonged occlusions.
These failures are mainly caused by delayed high-level belief updates or insufficient visual evidence.
For example, when an occluded agent appears abruptly between two slow reasoning updates, the controller may temporarily rely on an outdated belief context.
Future work will explore faster belief updates, adaptive hypothesis allocation, and stronger uncertainty calibration for safety-critical deployment.
\clearpage
\vspace{-6pt}
\end{document}